\documentclass[sigconf,nonacm]{acmart}

\usepackage{multirow}
\usepackage[utf8]{inputenc} % allow utf-8 input
\usepackage[T1]{fontenc}    % use 8-bit T1 fonts
\usepackage{url}            % simple URL typesetting
\usepackage{booktabs}       % professional-quality tables
\usepackage{amsfonts}       % blackboard math symbols
\usepackage{nicefrac}       % compact symbols for 1/2, etc.
\usepackage{microtype}      % microtypography
\usepackage{xcolor}         % colors
\usepackage{colortbl}       % table colors
\usepackage{float}
\usepackage{amsmath}
\usepackage{amsthm}
\usepackage{latexsym}
\usepackage{thmtools}
\usepackage{thm-restate}
\usepackage{etoolbox}
\usepackage{array}
\usepackage{booktabs}
\usepackage{makecell}
\usepackage{tabularx}
\usepackage{enumitem}
\usepackage{titlesec}
\usepackage{caption}
\usepackage{subcaption}
\usepackage{comment}
\usepackage{algorithm}
\usepackage{algorithmicx}
\usepackage[noend]{algpseudocode}
\usepackage{pifont}
\usepackage[framemethod=TikZ]{mdframed}
\usepackage{verbatim,ifthen,alltt}
\usepackage{microtype}      % microtypography
\usepackage{stmaryrd}
\usepackage{xfrac}
\usepackage{tikz}
\usepackage{pgf}
\usepackage{forest}

\usetikzlibrary{fit,calc,tikzmark}
\usetikzlibrary{arrows,decorations.pathmorphing,decorations.pathreplacing,decorations.footprints,fadings,calc,trees,mindmap,shadows,decorations.text,patterns,positioning,shapes}
\usetikzlibrary{arrows,shadows,backgrounds}
\usetikzlibrary{arrows.meta}
\usetikzlibrary{positioning,fit,automata,matrix}
\usetikzlibrary{shapes.symbols,shapes.misc,shapes.arrows}

\definecolor{darkred}{rgb}{0.7,0.1,0.1}
\definecolor{medred}{rgb}{0.5,0.1,0.1}
\definecolor{midred}{rgb}{0.7,0.2,0.2}
\definecolor{vdarkred}{rgb}{0.4,0.1,0.1}
\definecolor{darkslategray}{rgb}{0.18, 0.31, 0.31} %#2F4F4F
\definecolor{platinum}{rgb}{0.9, 0.89, 0.89} %#E5E4E2
\definecolor{gray}{rgb}{.4,.4,.4}
\definecolor{midgrey}{rgb}{0.5,0.5,0.5}
\definecolor{middarkgrey}{rgb}{0.35,0.35,0.35}
\definecolor{darkgrey}{rgb}{0.3,0.3,0.3}
\definecolor{darkred}{rgb}{0.7,0.1,0.1}
\definecolor{midblue}{rgb}{0.2,0.2,0.7}
\definecolor{darkblue}{rgb}{0.1,0.1,0.5}
\definecolor{darkgreen}{rgb}{0.1,0.5,0.1}
\definecolor{defseagreen}{cmyk}{0.69,0,0.50,0}
\definecolor{purple3}{RGB}{125,38,205}          % purple3

\definecolor{tyellow1}{HTML}{FCE94F}
\definecolor{tyellow2}{HTML}{EDD400}
\definecolor{tyellow3}{HTML}{C4A000}
\definecolor{torange1}{HTML}{FCAF3E}
\definecolor{torange2}{HTML}{F57900}
\definecolor{torange3}{HTML}{C35C00}
\definecolor{tbrown1}{HTML}{E9B96E}
\definecolor{tbrown2}{HTML}{C17D11}
\definecolor{tbrown3}{HTML}{8F5902}
\definecolor{tgreen1}{HTML}{8AE234}
\definecolor{tgreen2}{HTML}{73D216}
\definecolor{tgreen3}{HTML}{4E9A06}
\definecolor{tblue1}{HTML}{729FCF}
\definecolor{tblue2}{HTML}{3465A4}
\definecolor{tblue3}{HTML}{204A87}
\definecolor{tpurple1}{HTML}{AD7FA8}
\definecolor{tpurple2}{HTML}{75507B}
\definecolor{tpurple3}{HTML}{5C3566}
\definecolor{tred1}{HTML}{EF2929}
\definecolor{tred2}{HTML}{CC0000}
\definecolor{tred3}{HTML}{A40000}
\definecolor{tlgray1}{HTML}{EEEEEC}
\definecolor{tlgray2}{HTML}{D3D7CF}
\definecolor{tlgray3}{HTML}{BABDB6}
\definecolor{tdgray1}{HTML}{888A85}
\definecolor{tdgray2}{HTML}{555753}
\definecolor{tdgray3}{HTML}{2E3436}

\newcommand{\hlight}[1]{{\color{darkred}#1}}
\newcommand{\rhlight}[1]{\hlight{#1}}

\newcommand{\dbhlight}[1]{{\color{darkblue}#1}}

\newcommand{\dghlight}[1]{{\color[RGB]{0,120,0}#1}}
\newcommand{\nohlight}[1]{{#1}}

\usepackage[noabbrev,nameinlink,capitalise]{cleveref}
\AtBeginDocument{%
  
  \hypersetup{breaklinks=true}
}

\setcopyright{none}
\copyrightyear{2023}
\acmYear{2023}
\acmDOI{XXXXXXX.XXXXXXX}

\acmConference[arXiv preprint]{}{June 2023}{IRIT, Toulouse, France} %
\makeatletter
\def\thm@space@setup{\thm@preskip=5.0pt
\thm@postskip=0pt}
\makeatother

\newtheoremstyle{newstyle}      
{5.5pt} %Aboveskip 
{-2.0pt} %Below skip
{\mdseries} %Body font e.g.\mdseries,\bfseries,\scshape,\itshape
{} %Indent
{\bfseries} %Head font e.g.\bfseries,\scshape,\itshape
{.} %Punctuation afer theorem header
{ } %Space after theorem header
{} %Heading
\theoremstyle{newstyle}

\crefname{theorem}{Theorem}{Theorems}
\crefname{lemma}{Lemma}{Lemmas}
\crefname{proposition}{Proposition}{Propositions}
\crefname{definition}{Definition}{Definitions}
\crefname{corollary}{Corollary}{Corollaries}
\crefname{example}{Example}{Examples}
\crefname{claim}{Claim}{Claims}
\crefname{assumption}{Assumption}{Assumptions}
\crefname{enumi}{}{}



\newcommand{\fml}[1]{{\mathcal{#1}}}

\newcommand{\tn}[1]{\textnormal{#1}}

\newcommand{\mbf}[1]{\ensuremath\mathbf{#1}}
\newcommand{\msf}[1]{\ensuremath\mathsf{#1}}
\newcommand{\mbb}[1]{\ensuremath\mathbb{#1}}

\newcommand{\axp}{\ensuremath\mathsf{AXp}}
\newcommand{\cxp}{\ensuremath\mathsf{CXp}}

\newcommand{\sv}{\ensuremath\msf{Sv}}

\newcommand{\oper}[1]{\ensuremath\mathsf{#1}}

\newcommand{\bigland}{\ensuremath\bigwedge}

\newcommand{\relevant}{\oper{Relevant}}
\newcommand{\irrelevant}{\oper{Irrelevant}}

\newcounter{tableeqn}[table]

\DeclareMathOperator*{\limply}{\rightarrow}

\newcommand{\jnoteF}[1]{}

\newcolumntype{L}[1]{>{\raggedright\let\newline\\\arraybackslash\hspace{0pt}}m{#1}}
\newcolumntype{C}[1]{>{\centering\let\newline\\\arraybackslash\hspace{0pt}}m{#1}}
\newcolumntype{R}[1]{>{\raggedleft\let\newline\\\arraybackslash\hspace{0pt}}m{#1}}

\tikzset{
  0 my edge/.style={densely dashed, my edge, draw=midblue},
  my edge/.style={-{Stealth[]}, draw=midblue},
}

\def\HiLi{\leavevmode\rlap{\hbox to \linewidth{\color{platinum}\leaders\hrule height .8\baselineskip depth .5ex\hfill}}}

\titlespacing{\section}{0pt}{*2.15}{*1.0}
\titlespacing{\subsection}{0pt}{*1.25}{*0.75}
\titlespacing{\subsubsection}{0pt}{*0.35}{*0.5}
\titlespacing{\paragraph}{0pt}{*0.1}{*0.575}

\makeatletter
\newcommand\nparagraph{%
  \@startsection{paragraph}
    {4}
    {\z@}
    {0.225ex \@plus0.225ex \@minus.125ex}
    {-1em}
    {\normalfont\normalsize\bfseries}%
}
\makeatother

\setitemize{noitemsep,topsep=0pt,parsep=0pt,partopsep=0pt}
\setenumerate{noitemsep,topsep=0pt,parsep=0pt,partopsep=0pt}
\setlist{nosep,leftmargin=0.45cm}

\providetoggle{long}
\settoggle{long}{false}

\makeatletter
\algnewcommand{\LineComment}[1]{\Statex \hskip\ALG@thistlm \(\triangleright\) #1}
\makeatother

\hypersetup{breaklinks=true}

\begin{document}

\title{Explainability is NOT a Game -- Preliminary Report}
\titlenote{An extensively revised
  %comprehensively modified 
  version of this paper has been accepted for publication in
  the Communications of the ACM.}

\author{Joao Marques-Silva}
\authornote{Both authors contributed equally to this research.}
\email{joao.marques-silva@irit.fr}
\orcid{0000-0002-6632-3086}
\affiliation{%
  \institution{IRIT, CNRS}
  %\streetaddress{No Street}
  \city{Toulouse}
  %\state{No State}
  \country{France}
  \postcode{31400}
}

\author{Xuanxiang Huang}
\authornotemark[2]
\email{xuanxiang.huang@univ-toulouse.fr}
\orcid{0000-0002-3722-7191}
\affiliation{%
  \institution{University of Toulouse}
  %\streetaddress{No Street}
  \city{Toulouse}
  %\state{No State}
  \country{France}
  \postcode{31400}
}

%% \author{Lars Th{\o}rv{\"a}ld}
%% \affiliation{%
%%   \institution{The Th{\o}rv{\"a}ld Group}
%%   \streetaddress{1 Th{\o}rv{\"a}ld Circle}
%%   \city{Hekla}
%%   \country{Iceland}}
%% \email{larst@affiliation.org}

%% \author{Valerie B\'eranger}
%% \affiliation{%
%%   \institution{Inria Paris-Rocquencourt}
%%   \city{Rocquencourt}
%%   \country{France}
%% }

%% \author{Aparna Patel}
%% \affiliation{%
%%  \institution{Rajiv Gandhi University}
%%  \streetaddress{Rono-Hills}
%%  \city{Doimukh}
%%  \state{Arunachal Pradesh}
%%  \country{India}}

%% \author{Huifen Chan}
%% \affiliation{%
%%   \institution{Tsinghua University}
%%   \streetaddress{30 Shuangqing Rd}
%%   \city{Haidian Qu}
%%   \state{Beijing Shi}
%%   \country{China}}

%% \author{Charles Palmer}
%% \affiliation{%
%%   \institution{Palmer Research Laboratories}
%%   \streetaddress{8600 Datapoint Drive}
%%   \city{San Antonio}
%%   \state{Texas}
%%   \country{USA}
%%   \postcode{78229}}
%% \email{cpalmer@prl.com}

%% \author{John Smith}
%% \affiliation{%
%%   \institution{The Th{\o}rv{\"a}ld Group}
%%   \streetaddress{1 Th{\o}rv{\"a}ld Circle}
%%   \city{Hekla}
%%   \country{Iceland}}
%% \email{jsmith@affiliation.org}

%% \author{Julius P. Kumquat}
%% \affiliation{%
%%   \institution{The Kumquat Consortium}
%%   \city{New York}
%%   \country{USA}}
%% \email{jpkumquat@consortium.net}

\renewcommand{\shortauthors}{Marques-Silva and Huang}

\begin{abstract}
  Explainable artificial intelligence (XAI) aims to help human
  decision-makers in understanding complex machine learning (ML)
  models.
  One of the hallmarks of XAI are measures of relative feature
  importance, which are theoretically justified through the use of
  Shapley values.
  This paper builds on recent work and offers a simple argument for
  why Shapley values can provide misleading measures of relative
  feature importance, by assigning more importance to features that
  are irrelevant for a prediction, and assigning less importance to
  features that are relevant for a prediction.
  The significance of these results is that they effectively challenge
  the many proposed uses of measures of relative feature importance in
  a fast-growing range of high-stakes application domains.
\end{abstract}

\begin{CCSXML}
<ccs2012>
<concept>
<concept_id>10010147.10010178</concept_id>
<concept_desc>Computing methodologies~Artificial intelligence</concept_desc>
<concept_significance>500</concept_significance>
</concept>
<concept>
<concept_id>10010147.10010257.10010321</concept_id>
<concept_desc>Computing methodologies~Machine learning algorithms</concept_desc>
<concept_significance>500</concept_significance>
</concept>
<concept>
<concept_id>10003752.10003790.10003794</concept_id>
<concept_desc>Theory of computation~Automated reasoning</concept_desc>
<concept_significance>500</concept_significance>
</concept>
<concept>
<concept_id>10010147.10010257</concept_id>
<concept_desc>Computing methodologies~Machine learning</concept_desc>
<concept_significance>500</concept_significance>
</concept>
</ccs2012>
\end{CCSXML}

\ccsdesc[500]{Computing methodologies~Artificial intelligence}
\ccsdesc[500]{Computing methodologies~Machine learning algorithms}
\ccsdesc[500]{Theory of computation~Automated reasoning}
\ccsdesc[500]{Computing methodologies~Machine learning}

\keywords{Explainable AI, Shapley values, Abductive reasoning}

%\received{31 July 2023}
%\received[revised]{12 March 2023}
%\received[accepted]{5 June 2023}

\maketitle

\section{Introduction} \label{sec:intro}

The societal and economic significance of machine learning (ML) cannot
be overstated, with many remarkable advances made in recent years.
However, the operation of complex ML models is most often inscrutable,
with the consequence that decisions taken by ML models cannot be
fathomed by human decision makers.
It is therefore of importance to devise automated approaches to
explain the predictions made by complex ML models.
This is the main motivation for \dbhlight{eXplainable AI}
(\dbhlight{XAI}).
Explanations thus serve to build trust, but also to debug complex
systems of AI.
Furthermore, in situations where decisions of ML models impact people,
one should expect explanations to offer the strongest guarantees of
rigor. 

However, the most popular XAI
approaches~\cite{muller-plosone15,guestrin-kdd16,lundberg-nips17,guestrin-aaai18,xai-bk19}
offer no guarantees of rigor. Unsurprisingly, a number of works have
demonstrated several misconceptions of informal approaches to
XAI~\cite{inms-corr19,nsmims-sat19,ignatiev-ijcai20,iims-jair22,hms-corr23,ms-iceccs23}. 
%%
%%\jnote{What is wrong with informal XAI?}
%%\jnote{What is formal XAI?}
%%
In contrast to informal XAI, formal explainability offers a
logic-based, model-precise approach for computing
explanations~\cite{inms-aaai19}.
Although formal explainability also exhibits a number of drawbacks,
including the computational complexity of logic-based reasoning, there
has been continued progress since its
inception~\cite{,msi-aaai22,ms-rw22}.

%%\jnote{Intro next paragraph: among the informal approaches to XAI... }

Among the existing informal approaches to XAI, the use of Shapley
values as a mechanism for feature attribution is arguably the
best-known.
Shapley values~\cite{shapley-ctg53} were originally proposed in the
context of game theory, but have found a wealth of application
domains~\cite{roth-bk88}.
More importantly, for more than two decades Shapley values have been
proposed in the context of explaining the decisions of complex ML
models~\cite{conklin-asmbi01,kononenko-jmlr10,kononenko-kis14,lundberg-nips17}.
The importance of Shapley values for explainability is illustrated by
the massive impact of tools like SHAP~\cite{lundberg-nips17},
including many recent uses that have a direct influence on human
beings (see~\cite{hms-corr23} for some recent references).

Unfortunately, the exact computation of Shapley values in the case of
explainability has not been studied in practice, in part because
of its computational complexity. Hence, it is unclear how good
are existing approximate solutions, with a well-known example being
SHAP~\cite{lundberg-nips17,lundberg-naturemi20,lundberg-corr22}.
Recent work~\cite{barcelo-aaai21} proposed a polynomial-time algorithm
for computing Shapley values in the case of classifiers represented by
deterministic decomposable boolean circuits. As a result, and for one
concrete family of classifiers, it became possible to compare the
estimates of tools such as SHAP~\cite{lundberg-nips17} with those
obtained with exact algorithms.

Furthermore, since Shapley values aim to measure the relative
importance of features, a natural question is whether the relative
importance of features obtained with Shapley values can indeed be
trusted. Given that the definition of Shapley values is axiomatic, one
may naturally question how reliable those values are. Evidently, if 
the relative order of features dictated by Shapley values can be
proved inadequate, then the use of Shapley values in explainability 
ought to be deemed unworthy of trust.

A number of earlier works reported practical problems with
explainability approaches based on Shapley
values~\cite{friedler-icml20} (\cite{hms-corr23} covers a number of
additional references). However, these works focus on practical tools,
which approximate Shapley values, but do not investigate the possible
existence of fundamental limitations with the use of Shapley values in
explainability.
In contrast with these other works, this paper offers a simple
argument for why relative feature importance obtained with Shapley
values can provide misleading information, in that features that bear
no significance for a prediction can be deemed more important, in
terms of Shapley values, than features that bear some significance for
the same prediction.
The importance of this paper's results, and of the identified flaws
with Shapley values, should be assessed in light of the fast-growing
uses of explainability solutions in domains that directly impact human
beings, e.g.\ medical diagnostic applications, especially when the
vast majority of such uses build on Shapley values for
explainability.

The paper is organized as follows.
\cref{sec:prelim} introduces the notation and definitions used
throughout. This includes a brief introduction to formal explanations,
but also to Shapley values for explainability.
\cref{sec:xpr} revisits the concepts of relevancy/irrelevancy, which
have been studied in logic-based abduction since the mid
1990s~\cite{gottlob-jacm95}.
\cref{sec:issues} demonstrates the inadequacy of Shapley values for
feature attribution.
Finally,~\cref{sec:conc} discusses the paper's results, but it also
briefly examines additional flaws of Shapley values.

\jnoteF{Others have reported limitations with using Shapley values; we
  prove in this paper why that is the case, and why there are
  fundamental limitations to the use of Shapley values.}

\jnoteF{The paper offers a simple argument for why Shapley values will
  provide misleading information to human decision makers.}

%\section{Preliminaries} \label{sec:prelim}
\section{Definitions} \label{sec:prelim}

Throughout the paper, we adopt the notation and the definitions
introduced in earlier work, namely~\cite{msi-aaai22,ms-rw22}
and also~\cite{barcelo-aaai21}.

\subsection{Classification Problems} \label{ssec:clssf}
%\paragraph{\dbhlight{{\bfseries{Classification problems.}}}}

A classification problem is defined on a set of features
$\fml{F}=\{1,\ldots,m\}$, and a set of classes
$\fml{K}=\{c_1,\ldots,c_K\}$.
Each feature $i\in\fml{F}$ takes values from a domain $\fml{D}_i$.
Domains can be ordinal (e.g.\ real- or integer-valued) or
categorical. 
Feature space is defined by the cartesian product of the domains of
the features: $\mbb{F}=\fml{D}_1\times\cdots\times\fml{D}_m$.
A classifier $\fml{M}$ computes a (non-constant) classification
function: $\kappa:\mbb{F}\to\fml{K}$\footnote{%
A classifier that computes a constant function, i.e.\ the same
prediction for all points in feature space, is of course
uninteresting, and so it is explicitly disallowed.}.
A classifier $\fml{M}$ is associated with a tuple
$(\fml{F},\mbb{F},\fml{K},\kappa)$.
For the purposes of this paper, we restrict $\kappa$ to be a
non-constant boolean function. This restriction does not in any way
impact the validity of our results.

Given a classifier $\fml{M}$, and a point $\mbf{v}\in\mbb{F}$, with
$c=\kappa(\mbf{v})$ and $c\in\fml{K}$, $(\mbf{v},c)$ is referred to as
an \emph{instance} (or sample). An explanation problem $\fml{E}$ is
associated with a tuple $(\fml{M},(\mbf{v},c))$.
As a result, $\mbf{v}$ represents a concrete point in feature space,
whereas $\mbf{x}\in\mbb{F}$ represents an arbitrary point in feature
space.

\begin{figure*}[t]
  \begin{mdframed}[linewidth=1.5pt,linecolor=darkblue,roundcorner=5pt] %skipabove=10pt
    \begin{subfigure}[c]{0.5\textwidth}
      % Example from Rudin et al. NeurIPS'19 paper
%%
%\tikzset{every label/.style={xshift=-0.35ex,
%  yshift=-5.25ex,
%  text width=1ex,
%  align=right, inner sep=1pt, font=\tiny, text=midblue}}
%%
%\tikzset{tlabel/.style={xshift=0.25ex, yshift=1.75ex, text width=1ex,
%    align=right, inner sep=1pt, font=\tiny, text=midblue}}
%%%\tikzset{every node/.style={---rectangle---}}
%
\forestset{
  BDT/.style={
    for tree={
      l=1.5cm,s sep=1.15cm,
      if n children=0{}{circle}, %rectangle
      %if n children=0{}{draw},
      draw=midblue,%draw=black,%
      text=midblue,%text=black,%
      edge={
        my edge
      },
      %if n=1{
      %  edge+={0 my edge},
      %}{},
      edge=thick,
    }
  },
}
%
% middle-middle=x : x_1
% top-left=x : x_2
% bottom-right=x : x_3
% bottom-left=x : x_4
% top-right=x : x_5
%
\begin{forest}
  BDT
  [{$x_1$}, label={[yshift=-6.875ex]{{\tiny1}}} %middle-middle=x
    [{$x_2$}, edge={very thick}, label={[yshift=-6.875ex]{{\tiny2}}}, %top-left=x
      edge label={node[midway,left,xshift=-0.5pt] {{\scriptsize$=0$}}}
      [{$x_3$}, edge={very thick}, label={[yshift=-6.875ex]{{\tiny4}}}, %bottom-right=x
        edge label={node[midway,left,xshift=-2.5pt] {{\scriptsize$=0$}}}
        [{$x_4$}, edge={very thick}, label={[yshift=-6.875ex]{{\tiny6}}}, %bottom-left=x
          edge label={node[midway,left,xshift=0.5pt] {{\scriptsize$=0$}}}
          [\rhlight{\textbf{0}}, edge={very thick}, label={[yshift=-5.25ex]{{\tiny10}}},
            edge label={node[midway,left,xshift=-1.5pt] {{\scriptsize$=0$}}},
            rectangle, fill={tred3!20} ]
          [\dghlight{\textbf{1}}, label={[yshift=-5.25ex]{{\tiny11}}},
            edge label={node[midway,right,xshift=0.5pt] {{\scriptsize$=1$}}},
            rectangle, fill={tgreen3!25} ]
        ]
        [{$x_4$}, label={[yshift=-6.875ex]{{\tiny7}}}, %bottom-left=x
          edge label={node[midway,right,xshift=0.5pt] {{\scriptsize$=1$}}}
          [\dghlight{\textbf{1}}, label={[yshift=-5.25ex]{{\tiny12}}},
            edge label={node[midway,left,xshift=-1.5pt] {{\scriptsize$=0$}}},
            rectangle, fill={tgreen3!25} ]
          [\rhlight{\textbf{0}}, label={[yshift=-5.25ex]{{\tiny13}}},
            edge label={node[midway,right,xshift=0.5pt] {{\scriptsize$=1$}}},
            rectangle, fill={tred3!20} ]
        ]
      ]
      [{$x_3$}, label={[yshift=-6.875ex]{{\tiny5}}}, %bottom-left=x
        edge label={node[midway,right,xshift=1.5pt] {{\scriptsize$=1$}}}
        [\rhlight{\textbf{0}}, label={[yshift=-5.25ex]{{\tiny8}}},
          edge label={node[midway,left,xshift=0.5pt] {{\scriptsize$=1$}}},
          rectangle, fill={tred3!20} ]
        [{$x_4$}, label={[yshift=-6.875ex]{{\tiny9}}}, %top-right=x
          edge label={node[midway,right,xshift=-0.5pt] {{\scriptsize$=0$}}}
          [\rhlight{\textbf{0}}, label={[yshift=-5.25ex]{{\tiny14}}},
            edge label={node[midway,left,xshift=-0.5pt] {{\scriptsize$=1$}}},
            rectangle, fill={tred3!20} ]
          [\dghlight{\textbf{1}}, label={[yshift=-5.25ex]{{\tiny15}}},
            edge label={node[midway,right,xshift=0.5pt] {{\scriptsize$=0$}}},
            rectangle, fill={tgreen3!25} ]
        ]
      ]
    ]
    [\rhlight{\textbf{0}}, label={[yshift=-5.25ex]{{\tiny3}}},
      edge label={node[midway,right,xshift=0.5pt] {{\scriptsize$=1$}}},
      rectangle, fill={tred3!20} ]
  ]
\end{forest}
      %%%$\kappa(\cdot) = \ldots$
    \end{subfigure}
    \begin{subfigure}[c]{0.5\textwidth}
      \begin{center}
        \begin{tabular}{cccccc} \toprule
          row \# & $x_1$ & $x_2$ & $x_3$ & $x_4$ & $\kappa(\mbf{x})$
          \\ \toprule
          1 & 0 & 0 & 0 & 0 & 0 \\
          2 & 0 & 0 & 0 & 1 & 1 \\
          3 & 0 & 0 & 1 & 0 & 1 \\
          4 & 0 & 0 & 1 & 1 & 0 \\
          5 & 0 & 1 & 0 & 0 & 1 \\
          6 & 0 & 1 & 0 & 1 & 0 \\
          7 & 0 & 1 & 1 & 0 & 0 \\
          8 & 0 & 1 & 1 & 1 & 0 \\
          9 & 1 & 0 & 0 & 0 & 0 \\
          10 & 1 & 0 & 0 & 1 & 0 \\
          11 & 1 & 0 & 1 & 0 & 0 \\
          12 & 1 & 0 & 1 & 1 & 0 \\
          13 & 1 & 1 & 0 & 0 & 0 \\
          14 & 1 & 1 & 0 & 1 & 0 \\
          15 & 1 & 1 & 1 & 0 & 0 \\
          16 & 1 & 1 & 1 & 1 & 0 \\
          \bottomrule
        \end{tabular}
      \end{center}
    \end{subfigure}
    %\captionof{figure}{Example classifier -- decision tree} \label{fig:example01}
    \caption{Example classifier -- decision tree and its truth table.
     For this classifier, we have $\fml{F}=\{1,2,3,4\}$,
     $\fml{D}_i=\{0,1\},i=1,2,3,4$, $\mbb{F}=\{0,1\}^4$, and
     $\fml{K}=\{0,1\}$. The classification function is given by the
     decision tree shown, or alternatively by the truth table.
     Finally, the instance considered is $((0,0,0,0),0)$,
     corresponding to row 1 in the truth table. The instance is
     consistent with path $\langle1,2,4,6,10\rangle$, which is
     highlighted in the DT. The prediction is 0, as indicated in
     terminal node 10.
    }
    \label{fig:example01}
  \end{mdframed}
\end{figure*}

As a running example, we consider the decision tree (DT) shown 
in~\cref{fig:example01}. Since it will be used later, we also show the
truth table for the DT classifier.
Given the information shown in the DT, we have that
$\fml{F}=\{1,2,3,4\}$, $\fml{D}_i=\{0,1\},i=1,2,3,4$,
$\mbb{F}=\{0,1\}^4$, and $\fml{K}=\{0,1\}$. The classification
function $\kappa$ is given by the decision tree shown, or
alternatively by the truth table. Finally, the instance considered is
$(\mbf{v},c)=((0,0,0,0),0)$.

\subsection{Formal Explanations} \label{ssec:fxai}
%\paragraph{\dbhlight{{\bfseries{Formal explanations.}}}}

The presentation of formal explanations follows recent
accounts~\cite{ms-rw22}.
In the context of XAI, abductive explanations (AXp's) have been
studied since 2018~\cite{darwiche-ijcai18,inms-aaai19}. Similar to
other heuristic approaches, e.g.\ Anchors~\cite{guestrin-aaai18},
abductive explanations are an example of explainability by feature
selection, i.e.\ a subset of features is selected as the explanation.
AXp's represent a rigorous example of explainability by feature
selection, and can be viewed as the answer to a ``\emph{\rhlight{Why
  (the prediction)?}}'' question. 
An AXp is defined as a subset-minimal (or irreducible) set of features
$\fml{X}\subseteq\fml{F}$ such that the features in $\fml{X}$ are
sufficient for the prediction. 
This is to say that, if the features in $\fml{X}$ are fixed to the
values determined by $\mbf{v}$, then the prediction is guaranteed to
be $c=\kappa(\mbf{v})$.
The sufficiency for the prediction can be stated formally:
\begin{equation} \label{eq:waxp}
  \forall(\mbf{x}\in\mbb{F}).\left[\bigland\nolimits_{i\in\fml{X}}(x_i=v_i)\right]\limply(\kappa(\mbf{x})=\kappa(\mbf{v}))
\end{equation}

Observe that \eqref{eq:waxp} is monotone on $\fml{X}$, and so the two
conditions for a set $\fml{X}\subseteq\fml{F}$ to be an AXp
(i.e.\ sufficiency for prediction and subset-minimality), can be
stated as follows:
\begin{align} \label{eq:axp}
  &\forall(\mbf{x}\in\mbb{F}).\left[\bigland\nolimits_{i\in\fml{X}}(x_i=v_i)\right]\limply(\kappa(\mbf{x})=\kappa(\mbf{v}))\land\\
  &
  \forall(t\in\fml{X}).\exists(\mbf{x}\in\mbb{F}).\left[\bigland\nolimits_{i\in\fml{X}\setminus\{t\}}(x_i=v_i)\right]\land(\kappa(\mbf{x})\not=\kappa(\mbf{v}))
  \nonumber
  %\\
\end{align}
A predicate $\axp:2^{\fml{F}}\to\{0,1\}$ is associated with
\eqref{eq:axp}, such that $\axp(\fml{X};\fml{E})$ holds true if and
only if \eqref{eq:axp} holds true\footnote{%
When defining concepts, we will show the necessary parameterizations.
However, in later uses, those parameterizations will be omitted, for
simplicity.}.

An AXp can be interpreted as a logic rule of the form:
\begin{equation}
  \tn{IF} \quad \left[\bigland\nolimits_{i\in\fml{X}}(x_i=v_i)\right] \quad
  \tn{THEN} \quad (\kappa(\mbf{x})=c) %%\kappa(\mbf{v})
\end{equation}
where $c=\kappa(\mbf{v})$.
It should be noted that informal XAI methods have also proposed the
use of IF-THEN rules~\cite{guestrin-aaai18} which, in the case of
Anchors~\cite{guestrin-aaai18} may or may not be
sound~\cite{inms-aaai19,ignatiev-ijcai20}. In contrast, rules obtained
from AXp's are logically sound.

Moreover, contrastive explanations (CXp's) represent a type of
explanation that differs from AXp's, in that CXp's answer a
``\emph{\rhlight{Why Not (some other prediction)?}}''
question~\cite{miller-aij19,inams-aiia20}.
Given a set $\fml{Y}\subseteq\fml{F}$, sufficiency for changing the
prediction can be stated formally:
\begin{equation} \label{eq:wcxp}
  \exists(\mbf{x}\in\mbb{F}).\left[\bigland\nolimits_{i\in\fml{F}\setminus\fml{Y}}(x_i=v_i)\right]\land(\kappa(\mbf{x})\not=\kappa(\mbf{v}))
\end{equation}
A CXp is a subset-minimal set of features which, if allowed to take a
value other than the value determined by $\mbf{v}$, then the
predicted can be changed by choosing suitable values to those features.

Similarly to the case of AXp's, for CXp's \eqref{eq:wcxp} is monotone
on $\fml{Y}$, and so the two conditions (sufficiency for changing the
prediction and subset-minimality) can be stated formally as follows:
\begin{align} \label{eq:cxp}
  &\exists(\mbf{x}\in\mbb{F}).\left[\bigland\nolimits_{i\in\fml{F}\setminus\fml{Y}}(x_i=v_i)\right]\land(\kappa(\mbf{x})\not=\kappa(\mbf{v}))\land\\
  &
  \forall(t\in\fml{Y}).\forall(\mbf{x}\in\mbb{F}).\left[\bigland\nolimits_{i\in\fml{F}\setminus(\fml{Y}\setminus\{t\})}(x_i=v_i)\right]\limply(\kappa(\mbf{x})=\kappa(\mbf{v}))
  \nonumber
  %\\
\end{align}
A predicate $\cxp:2^{\fml{F}}\to\{0,1\}$ is associated with
\eqref{eq:cxp}, such that $\cxp(\fml{Y};\fml{E})$ holds true if and
only if \eqref{eq:cxp} holds true.

Algorithms for computing AXp's and CXp's for different families of
classifiers have been proposed in recent years (\cite{msi-aaai22}
provides a recent account of the progress observed in computing formal
explanations). These algorithms include the use of automated reasoners
(e.g.\ SAT, SMT or MILP solvers), or dedicated algorithms for families
of classifiers for which computing one explanation is tractable.

Given an explanation problem $\fml{E}$, the sets of AXp's and CXp's
are represented by:
\begin{align}
  \mbb{A}(\fml{E}) = \{\fml{X}\subseteq\fml{F}\,|\,\axp(\fml{X};\fml{E})\}
  \label{eq:seta}
  \\
  \mbb{C}(\fml{E}) = \{\fml{Y}\subseteq\fml{F}\,|\,\cxp(\fml{Y};\fml{E})\}%\\
  \label{eq:setc}
\end{align}
For example, $\mbb{A}(\fml{E})$ represents the set of \dbhlight{all}
logic rules that predict $c=\kappa(\mbf{v})$, which are consistent
with $\mbf{v}$, and which are irreducible (i.e.\ no literal $x_i=v_i$
can be discarded).

Furthermore, it has been proved~\cite{inams-aiia20} that (i) a set
$\fml{X}\subseteq\fml{F}$ is an AXp if and only if it is a minimal
hitting set (MHS) of the set of CXp's; and (ii) a set
$\fml{Y}\subseteq\fml{F}$ is a CXp if and only if it is an MHS of the
set of AXp's.
This property is referred to as MHS duality, and can be traced back to
the seminal work of R.~Reiter~\cite{reiter-aij87} in model-based
diagnosis.
Moreover, MHS duality has been shown to be instrumental for the
enumeration of AXp's and CXp's, but also for answering other
explainability queries~\cite{ms-rw22}.

For the running example, and since it is feasible to represent the
function with a truth table, then there exist polyno\-mial-time
algorithms (on the size of the truth-table) for computing all AXp's
and all CXp's~\cite{hms-corr23}. This is illustrated
in~\cref{fig:xps}.
\cref{tab:ex01} illustrates how each set is analyzed when computing
AXp's or CXp's.

\begin{figure*}[t]
  \begin{mdframed}[linewidth=1.5pt,linecolor=darkblue,roundcorner=5pt]
    %skipabove=10pt
    %

    \medskip\smallskip
    \begin{center}
      \renewcommand{\tabcolsep}{0.5em}
      \begin{tabular}{cccc|cccc}
        \toprule[1pt]
        $\fml{S}$ &
        rows picked by $\fml{S}$ & $\kappa(\mbf{x})=c$? &
        $\fml{S}$ is AXp? &
        %% over $\fml{S}$
        $\fml{F}\setminus\fml{S}$ &
        rows picked by $\fml{F}\setminus\fml{S}$ &
        $\kappa(\mbf{x})\not=c$? &
        $\fml{S}$ is CXp?
        %% over $\fml{F}\setminus\fml{S}$
        \\
        \midrule[0.875pt]
        $\emptyset$ & 1..16 & No & &
        $\{1,2,3,4\}$ & 1 & No & 
        \\
        $\{1\}$ & 1,2,3,4,5,6,7,8 & No & &
        $\{2,3,4\}$ & 1,9 & No & 
        \\
        $\{2\}$ & 1,2,3,4,9,10,11,12 & No & &
        $\{1,3,4\}$ & 1,5 & Yes & Yes
        \\
        $\{3\}$ & 1,2,5,6,9,10,13,14 & No & &
        $\{1,2,4\}$ & 1,3 & Yes & Yes
        \\
        $\{4\}$ & 1,3,5,7,9,11,13,15 & No & & 
        $\{1,2,3\}$ & 1,2 & Yes & Yes
        \\
        $\{1,2\}$ & 1,2,3,4 & No & & 
        $\{3,4\}$ & 1,5,9,13 & Yes &
        \\
        $\{1,3\}$ & 1,2,5,6 & No & & 
        $\{2,4\}$ & 1,3,9,11 & Yes &
        \\
        $\{1,4\}$ & 1,3,5,7 & No & & 
        $\{2,3\}$ & 1,2,9,10 & Yes & 
        \\
        $\{2,3\}$ & 1,2,9,10 & No & & 
        $\{1,4\}$ & 1,3,5,7 & Yes & 
        \\
        $\{2,4\}$ & 1,3,9,11 & No & & 
        $\{1,3\}$ & 1,2,5,6 & Yes & 
        \\
        $\{3,4\}$ & 1,5,9,13 & No & & 
        $\{1,2\}$ & 1,2,3,4 & Yes & 
        \\
        $\{1,2,3\}$ & 1,2 & No & & 
        $\{4\}$ & 1,3,5,7,9,11,13,15 & Yes & 
        \\
        $\{1,2,4\}$ & 1,3 & No & & 
        $\{3\}$ & 1,2,5,6,9,10,13,14 & Yes & 
        \\
        $\{1,3,4\}$ & 1,5 & No & & 
        $\{2\}$ & 1,2,3,4,9,10,11,12 & Yes &  %1,2,3,4,9,10,11,12
        \\
        $\{2,3,4\}$ & 1,9 & Yes & Yes & 
        $\{1\}$ & 1,2,3,4,5,6,7,8 & Yes & %1,2,3,4,5,6,7,8
        \\
        $\{1,2,3,4\}$ & 1 & Yes & & 
        $\emptyset$ & 1..16 & Yes & 
        \\
        \bottomrule[1pt]
      \end{tabular}
    \end{center}
    \caption{Computing AXp's/CXp's for the example DT and instance
      $((0,0,0,0),0)$. All subsets of features are considered.
      For computing AXp's, and for some set $\fml{S}$, the features in
      $\fml{S}$ are fixed to their values as dictated by $\mbf{v}$.
      The picked rows are the rows consistent with those fixed
      values. For example, if $\fml{S}=\{2,3,4\}$, then only rows 1
      and 9 are consistent with having features 2, 3 and 4 assigned
      value 0.
      Similarly, for computing CXp's, and for some set $\fml{S}$, the
      features in $\fml{F}\setminus\fml{S}$ are fixed to their values
      as dictated by $\mbf{v}$. The picked rows are again the rows
      consistent with those fixed values. For example, if
      $\fml{S}=\{2\}$, then $\fml{F}\setminus\fml{S}=\{1,3,4\}$, and
      so only rows 1 and 5 are consistent with having features 1, 3
      and 4 assigned value 0.
      An AXp is an irreducible set of features that is sufficient for
      the prediction. In this example, only $\{2,3,4\}$ respects the
      criteria.
      Moreover, a CXp is an irreducible set of features which, if
      allowed to take any value from their domain, the prediction
      changes. For this example, $\{2\}$, $\{3\}$ and $\{4\}$ respect
      the criteria, i.e.\ by only changing one of these features, we
      are able to change the prediction.
    }
    \label{fig:xps}
  \end{mdframed}
\end{figure*}

\begin{table}[t]
  \begin{tabular}{cccccc} \toprule[1.0pt]
    $\fml{S}$ & Template & Rows & Points & Value & $\kappa(\mbf{x})=c$?
    \\ \toprule[1.0pt]
    \multirow{4}{*}{$\{2,3\}$} & \multirow{4}{*}{$(x_1,0,0,x_4)$}
    & 1  & $(0,0,0,0)$ & 0 & \multirow{4}{*}{No}\\ 
    & & 2  & $(0,0,0,1)$ & 1 & \\ 
    & & 9  & $(1,0,0,0)$ & 0 & \\ 
    & & 10 & $(1,0,0,1)$ & 0 & \\ \midrule[0.75pt]
    \multirow{2}{*}{$\{1,2,4\}$} & \multirow{2}{*}{$(0,0,x_3,0)$}
    & 1  & $(0,0,0,0)$ & 0 & \multirow{2}{*}{No}\\ 
    & & 3  & $(0,0,1,0)$ & 1 & \\  \midrule[0.75pt]
    \multirow{2}{*}{$\{2,3,4\}$} & \multirow{2}{*}{$(x_1,0,0,0)$}
    & 1  & $(0,0,0,0)$ & 0 & \multirow{2}{*}{Yes}\\ 
    & & 9  & $(1,0,0,0)$ & 0 & \\  %\midrule[0.75pt]
    \bottomrule[1.0pt]
  \end{tabular}
  \caption{Examples of how each set is analyzed when computing AXp's. 
    For CXp's, a similar approach is used. } \label{tab:ex01}
\end{table}

\jnoteF{Mention many other topics of FXAI, and cite surveys.}

Formal explainability has made significant progress in recent years,
covering a wide range of topics of research. \cite{ms-rw22} represents
a recent overview of the progress in the emerging field of formal
explainability.

\subsection{Shapley Values in Explainability} \label{ssec:svs}
%\paragraph{\dbhlight{{\bfseries{Shapley values in explainability.}}}}
%
Shapley values were proposed in the 1950s, in the context of game
theory~\cite{shapley-ctg53}, and find a wealth of
uses~\cite{roth-bk88}.
More recently, Shapley values have been extensively used for
explaining the predictions of ML models,
e.g.~\cite{conklin-asmbi01,kononenko-jmlr10,kononenko-kis14,zick-sp16,lundberg-nips17,jordan-iclr19,taly-cdmake20,lakkaraju-nips21,watson-facct22},
among a vast number of recent examples (see~\cite{hms-corr23} for a
more comprehensive list of references).
Shapley values represent one example of explainability by feature
attribution, i.e.\ some score is assigned to each feature as a form of
explanation.
The complexity of computing Shapley values (as proposed in
SHAP~\cite{lundberg-nips17}) has been studied in recent
years~\cite{barcelo-aaai21,vandenbroeck-aaai21,barcelo-corr21,vandenbroeck-jair22}.
This section provides a brief overview of how Shapley values for
explainability are computed.
Throughout, we build on the notation used in recent
work~\cite{barcelo-aaai21,barcelo-corr21}, which builds on the work 
of~\cite{lundberg-nips17}.

Let $\Upsilon:2^{\fml{F}}\to2^{\mbb{F}}$ be defined by,
%
%\footnote{%
%When defining concepts, we will show the necessary parameterizations.
%However, in later uses, those parameterizations will be omitted, for
%simplicity.},
%
\begin{equation} \label{eq:upsilon}
  \Upsilon(\fml{S};\mbf{v})=\{\mbf{x}\in\mbb{F}\,|\,\land_{i\in\fml{S}}x_i=v_i\}
\end{equation}
i.e.\ for a given set $\fml{S}$ of features, and parameterized by
the point $\mbf{v}$ in feature space, $\Upsilon(\fml{S};\mbf{v})$
denotes all the points in feature space that have in common with
$\mbf{v}$ the values of the features specified by $\fml{S}$.
Observe that $\Upsilon$ is also used (implicitly) for picking the set
of rows where are interested in when computing explanations
(see~\cref{tab:ex01}).

Also, let $\phi:2^{\fml{F}}\to\mbb{R}$ be defined by,
\begin{equation} \label{eq:phi}
  \phi(\fml{S};\fml{M},\mbf{v})=\frac{1}{2^{|\fml{F}\setminus\fml{S}|}}\sum\nolimits_{\mbf{x}\in\Upsilon(\fml{S};\mbf{v})}\kappa(\mbf{x})
\end{equation}
Thus, given a set $\fml{S}$ of features,
$\phi(\fml{S};\fml{M},\mbf{v})$ represents the average value of the
classifier over the points of feature space represented by
$\Upsilon(\fml{S};\mbf{v})$.
%
%Clearly, $\phi$ gives the average (expected) value of the classifier
%in the point given by $\Upsilon(\fml{S}$.
%
The formulation presented in earlier
work~\cite{barcelo-aaai21,barcelo-corr21} allows for different input
distributions when computing the average values. For the purposes of
this paper, it suffices to consider solely a uniform input
distribution, and so the dependency on the input distribution is not
accounted for. 
%However, assuming a uniform distribution suffices for the purposes of
%this paper.
%

%
%(A distribution different from uniform could be considered, but it
%would not change the paper's conclusions.)
%
\begin{table}[t]
  \begin{tabular}{ccccc} \toprule[1.0pt]
    $\fml{S}$ & Template & $\Upsilon(\fml{S};\mbf{v})$ & Rows & 
    $\phi(\fml{S})$
    \\ \toprule[1.0pt]
    $\{1,4\}$ & $(0,x_2,x_3,0)$ &
    $\begin{array}{c}\{(0,0,0,0),\\(0,0,1,0),\\(0,1,0,0),\\(0,1,1,0)\}\end{array}$ &
    $\{1,3,5,7\}$ &
    $\sfrac{2}{4}$
    \\ \midrule[0.75pt]
    $\{3,4\}$ & $(x_1,x_2,0,0)$ &
    $\begin{array}{c}\{(0,0,0,0),\\(0,1,0,0),\\(1,0,0,0),\\(1,1,0,0)\}\end{array}$ &
    $\{1,5,9,13\}$ &
    $\sfrac{1}{4}$
    \\ %\midrule[0.75pt]
    \bottomrule[1.0pt]
  \end{tabular}
  \caption{Computation of average values. Rows represents the numbers
    of rows to consider when computing the average value.
  } \label{tab:ex02}
\end{table}
\cref{tab:ex02} illustrates how the average value is computed for two
concrete sets of features.
For example, if $\fml{S}=\{1,4\}$, then features 1 and 4 are fixed to
value 0 (as dictated by $\mbf{v}$). We then allow all possible
assignments to features 2 and 3, obtaining
$\Upsilon(\{1,4\})=\{(0,0,0,0),(0,0,1,0),(0,1,0,0),(0,1,1,0)\}$.
To compute $\phi(\fml{S})$, we sum up the values of the rows of the
truth table indicated by $\Upsilon(\fml{S})$, and divide by the total
number of points, which is 4 in this case.

To simplify the notation, the following definitions are used
throughout,
\begin{align}
  \Delta(i, \fml{S}; \fml{M},\mbf{v}) & =
  \left( \phi(\fml{S}\cup\{i\};\fml{M},\mbf{v}) -
  \phi(\fml{S};\fml{M},\mbf{v}) \right) \\  
  \varsigma(\fml{S};\fml{M},\mbf{v}) & =
  \sfrac{|\fml{S}|!(|\fml{F}|-|\fml{S}|-1)!}{|\fml{F}|!}
\end{align}

Finally, let $\sv:\fml{F}\to\mbb{R}$, i.e.\ the Shapley value for
feature $i$, be defined by,
%\footnote{%
%We distinguish $\shap(\cdot;\cdot,\cdot)$ from $\sv(\cdot;\cdot,\cdot)$. Whereas
%$\shap(\cdot;\cdot,\cdot)$ represents the value computed by the tool
%SHAP~\cite{lundberg-nips17}, $\sv(\cdot;\cdot,\cdot)$ represents the Shapley
%value in the context of (feature attribution based) explainability, as
%studied in a number of
%works~\cite{kononenko-jmlr10,kononenko-kis14,lundberg-nips17,barcelo-aaai21,vandenbroeck-aaai21,vandenbroeck-jair22}. Thus, 
%$\shap(\cdot;\cdot,\cdot)$ is a heuristic approximation of $\sv(\cdot;\cdot,\cdot)$.},
%
\begin{equation} \label{eq:sv}
  %\sv(i;\fml{M},\mbf{v})=\sum_{\fml{S}\subseteq(\fml{F}\setminus\{i\})}\frac{|\fml{S}|!(|\fml{F}|-|\fml{S}|-1)!}{|\fml{F}|!}\left(\phi(\fml{S}\cup\{i\};\fml{M},\mbf{v})-\phi(\fml{S};\fml{M},\mbf{v})\right)
  \sv(i;\fml{M},\mbf{v})=\sum\nolimits_{\fml{S}\subseteq(\fml{F}\setminus\{i\})}\varsigma(\fml{S};\fml{M},\mbf{v})\times\Delta(i,\fml{S};\fml{M},\mbf{v})
\end{equation}
Given an instance $(\mbf{v},c)$, the Shapley value assigned to each
feature measures the \emph{contribution} of that feature with respect
to the prediction. 
A positive/negative value indicates that the feature can contribute to
changing the prediction, whereas a value of 0 indicates no
contribution.

\begin{figure*}[t]
  \begin{mdframed}[linewidth=1.5pt,linecolor=darkblue,roundcorner=5pt]
    %skipabove=10pt
    %

    \medskip\smallskip
    %
    %% Feature 1:
    \begin{subfigure}{1.0\textwidth}
      \begin{center}
        \renewcommand{\tabcolsep}{0.5em}
        \begin{tabular}{cccccccc}
          \toprule[1pt]
          $\fml{S}$ & rows for $\fml{S}$ & rows for $\fml{S}\cup\{1\}$ &
          $\phi(\fml{S})$ & $\phi(\fml{S}\cup\{1\})$ & $\Delta(\fml{S})$ &
          $\varsigma(\fml{S})$ & $\varsigma(\fml{S})\times\Delta(\fml{S})$
          \\
          \midrule[0.875pt]
          $\emptyset$ & 1..16 & 1..8 & $\sfrac{3}{16}$ & $\sfrac{3}{8}$ &
          $\sfrac{3}{16}$ &
          $\sfrac{0!(4-1)!}{4!}=\sfrac{1}{4}$ & $\sfrac{3}{64}$
          \\
          $\{2\}$ & 1,2,3,4,9,10,11,12 & 1,2,3,4 &
          $\sfrac{2}{8}$ & $\sfrac{2}{4}$ & $\sfrac{1}{4}$ &
          $\sfrac{1!(4-2)!}{4!}=\sfrac{1}{12}$ & $\sfrac{1}{48}$
          \\
          $\{3\}$ & 1,2,5,6,9,10,13,14 & 1,2,5,6 & 
          $\sfrac{2}{8}$ & $\sfrac{2}{4}$ & $\sfrac{1}{4}$ &
          $\sfrac{1!(4-2)!}{4!}=\sfrac{1}{12}$ & $\sfrac{1}{48}$
          \\
          $\{4\}$ & 1,3,5,7,9,11,13,15 & 1,3,5,7 & 
          $\sfrac{2}{8}$ & $\sfrac{2}{4}$ & $\sfrac{1}{4}$ &
          $\sfrac{1!(4-2)!}{4!}=\sfrac{1}{12}$ & $\sfrac{1}{48}$
          \\
          $\{2,3\}$ & 1,2,9,10 & 1,2 & 
          $\sfrac{1}{4}$ & $\sfrac{1}{2}$ & $\sfrac{1}{4}$ &
          $\sfrac{2!(4-3)!}{4!}=\sfrac{1}{12}$ & $\sfrac{1}{48}$
          \\
          $\{2,4\}$ & 1,3,9,11 & 1,3 & 
          $\sfrac{1}{4}$ & $\sfrac{1}{2}$ & $\sfrac{1}{4}$ &
          $\sfrac{2!(4-3)!}{4!}=\sfrac{1}{12}$ & $\sfrac{1}{48}$
          \\
          $\{3,4\}$ & 1,5,9,13 & 1,5 &
          $\sfrac{1}{4}$ & $\sfrac{1}{2}$ & $\sfrac{1}{4}$ &
          $\sfrac{2!(4-3)!}{4!}=\sfrac{1}{12}$ & $\sfrac{1}{48}$
          \\
          $\{2,3,4\}$ & 1,9 & 1 & 
          0 & 0 & 0 &
          $\sfrac{3!(4-4)!}{4!}=\sfrac{1}{4}$ & 0
          \\
          \midrule[0.75pt]
          \multicolumn{7}{r}{Shapley value for feature 1 \hfill $\sv(1)~~=$} & $0.1719$ \\
          \bottomrule[1pt]
        \end{tabular}
      \end{center}
    \end{subfigure}
    %%\captionof{table}{Shapley value for feature 1} \label{tab:sv1}

    \medskip\medskip\medskip
    %
    %% Feature 2:
    \begin{subfigure}{1.0\textwidth}
      \begin{center}
        \renewcommand{\tabcolsep}{0.5em}
        \begin{tabular}{cccccccc}
          \toprule[1pt]
          $\fml{S}$ & rows for $\fml{S}$ & rows for $\fml{S}\cup\{2\}$ &
          $\phi(\fml{S})$ & $\phi(\fml{S}\cup\{2\})$ & $\Delta(\fml{S})$ &
          $\varsigma(\fml{S})$ & $\varsigma(\fml{S})\times\Delta(\fml{S})$
          \\
          \midrule[0.875pt]
          $\emptyset$ & 1..16 & 1,2,3,4,9,10,11,12 &
          $\sfrac{3}{16}$ & $\sfrac{2}{8}$ & $\sfrac{1}{16}$ &
          $\sfrac{0!(4-1)!}{4!}=\sfrac{1}{4}$ & $\sfrac{1}{64}$
          \\
          $\{1\}$ & 1..8 & 1,2,3,4 &
          $\sfrac{3}{8}$ & $\sfrac{2}{4}$ & $\sfrac{1}{8}$ &
          $\sfrac{1!(4-2)!}{4!}=\sfrac{1}{12}$ & $\sfrac{1}{96}$
          \\
          $\{3\}$ & 1,2,5,6,9,10,13,14 & 1,2,9,10 & 
          $\sfrac{2}{8}$ & $\sfrac{1}{4}$ & 0 &
          $\sfrac{1!(4-2)!}{4!}=\sfrac{1}{12}$ & 0
          \\
          $\{4\}$ & 1,3,5,7,9,11,13,15 & 1,3,9,11 & 
          $\sfrac{2}{8}$ & $\sfrac{1}{4}$ & 0 &
          $\sfrac{1!(4-2)!}{4!}=\sfrac{1}{12}$ & 0
          \\
          $\{1,3\}$ & 1,2,5,6 & 1,2 & 
          $\sfrac{2}{4}$ & $\sfrac{1}{2}$ & 0 &
          $\sfrac{2!(4-3)!}{4!}=\sfrac{1}{12}$ & 0
          \\
          $\{1,4\}$ & 1,3,5,7 & 1,3 & 
          $\sfrac{2}{4}$ & $\sfrac{1}{2}$ & 0 &
          $\sfrac{2!(4-3)!}{4!}=\sfrac{1}{12}$ & 0
          \\
          $\{3,4\}$ & 1,5,9,13 & 1,9 &
          $\sfrac{1}{4}$ & 0 & $-\sfrac{1}{4}$ &
          $\sfrac{2!(4-3)!}{4!}=\sfrac{1}{12}$ & $-\sfrac{1}{48}$
          \\
          $\{1,3,4\}$ & 1,5 & 1 & 
          $\sfrac{1}{2}$ & 0 & $-\sfrac{1}{2}$ &
          $\sfrac{3!(4-4)!}{4!}=\sfrac{1}{4}$ & $-\sfrac{1}{8}$
          \\
          \midrule[0.75pt]
          \multicolumn{7}{r}{Shapley value for feature 2 \hfill $\sv(2)~~=$} & $-0.1198$ \\
          \bottomrule[1pt]
        \end{tabular}
      \end{center}
      %%\captionof{table}{Shapley value for feature 2} \label{tab:sv2}
    \end{subfigure}    

    \medskip\medskip\medskip
    %
    %% Feature 3:
    \begin{subfigure}{1.0\textwidth}
      \begin{center}
        \renewcommand{\tabcolsep}{0.5em}
        \begin{tabular}{cccccccc}
          \toprule[1pt]
          $\fml{S}$ & rows for $\fml{S}$ & rows for $\fml{S}\cup\{3\}$ &
          $\phi(\fml{S})$ & $\phi(\fml{S}\cup\{3\})$ & $\Delta(\fml{S})$ &
          $\varsigma(\fml{S})$ & $\varsigma(\fml{S})\times\Delta(\fml{S})$
          \\
          \midrule[0.875pt]
          $\emptyset$ & 1..16 & 1,2,5,6,9,10,13,14 &
          $\sfrac{3}{16}$ & $\sfrac{2}{8}$ & $\sfrac{1}{16}$ &
          $\sfrac{0!(4-1)!}{4!}=\sfrac{1}{4}$ & $\sfrac{1}{64}$
          \\
          $\{1\}$ & 1..8 & 1,2,5,6 &
          $\sfrac{3}{8}$ & $\sfrac{2}{4}$ & $\sfrac{1}{8}$ &
          $\sfrac{1!(4-2)!}{4!}=\sfrac{1}{12}$ & $\sfrac{1}{96}$
          \\
          $\{2\}$ & 1,2,3,4,9,10,11,12 & 1,2,9,10 & 
          $\sfrac{2}{8}$ & $\sfrac{1}{4}$ & 0 &
          $\sfrac{1!(4-2)!}{4!}=\sfrac{1}{12}$ & 0
          \\
          $\{4\}$ & 1,3,5,7,9,11,13,15 & 1,5,9,13 & 
          $\sfrac{2}{8}$ & $\sfrac{1}{4}$ & 0 &
          $\sfrac{1!(4-2)!}{4!}=\sfrac{1}{12}$ & 0
          \\
          $\{1,2\}$ & 1,2,3,4 & 1,2 & 
          $\sfrac{2}{4}$ & $\sfrac{1}{2}$ & 0 &
          $\sfrac{2!(4-3)!}{4!}=\sfrac{1}{12}$ & 0
          \\
          $\{1,4\}$ & 1,3,5,7 & 1,5 & 
          $\sfrac{2}{4}$ & $\sfrac{1}{2}$ & 0 &
          $\sfrac{2!(4-3)!}{4!}=\sfrac{1}{12}$ & 0
          \\
          $\{2,4\}$ & 1,3,9,11 & 1,9 &
          $\sfrac{1}{4}$ & 0 & $-\sfrac{1}{4}$ &
          $\sfrac{2!(4-3)!}{4!}=\sfrac{1}{12}$ & $-\sfrac{1}{48}$
          \\
          $\{1,2,4\}$ & 1,3 & 1 & 
          $\sfrac{1}{2}$ & 0 & $-\sfrac{1}{2}$ &
          $\sfrac{3!(4-4)!}{4!}=\sfrac{1}{4}$ & $-\sfrac{1}{8}$
          \\
          \midrule[0.75pt]
          \multicolumn{7}{r}{Shapley value for feature 3 \hfill $\sv(3)~~=$} & $-0.1198$ \\
          \bottomrule[1pt]
        \end{tabular}
      \end{center}
      %%\captionof{table}{Shapley value for feature 2} \label{tab:sv3}
    \end{subfigure}    

    \medskip\medskip\medskip
    %
    %% Feature 4:
    \begin{subfigure}{1.0\textwidth}
      \begin{center}
        \renewcommand{\tabcolsep}{0.5em}
        \begin{tabular}{cccccccc}
          \toprule[1pt]
          $\fml{S}$ & rows for $\fml{S}$ & rows for $\fml{S}\cup\{4\}$ &
          $\phi(\fml{S})$ & $\phi(\fml{S}\cup\{4\})$ & $\Delta(\fml{S})$ &
          $\varsigma(\fml{S})$ & $\varsigma(\fml{S})\times\Delta(\fml{S})$
          \\
          \midrule[0.875pt]
          $\emptyset$ & 1..16 & 1,3,5,7,9,11,13,15 &
          $\sfrac{3}{16}$ & $\sfrac{2}{8}$ & $\sfrac{1}{16}$ &
          $\sfrac{0!(4-1)!}{4!}=\sfrac{1}{4}$ & $\sfrac{1}{64}$
          \\
          $\{1\}$ & 1..8 & 1,3,5,7 &
          $\sfrac{3}{8}$ & $\sfrac{2}{4}$ & $\sfrac{1}{8}$ &
          $\sfrac{1!(4-2)!}{4!}=\sfrac{1}{12}$ & $\sfrac{1}{96}$
          \\
          $\{2\}$ & 1,2,3,4,9,10,11,12 & 1,3,9,11 & 
          $\sfrac{2}{8}$ & $\sfrac{1}{4}$ & 0 &
          $\sfrac{1!(4-2)!}{4!}=\sfrac{1}{12}$ & 0
          \\
          $\{3\}$ & 1,2,5,6,9,10,13,14 & 1,5,9,13 & 
          $\sfrac{2}{8}$ & $\sfrac{1}{4}$ & 0 &
          $\sfrac{1!(4-2)!}{4!}=\sfrac{1}{12}$ & 0
          \\
          $\{1,2\}$ & 1,2,3,4 & 1,3 & 
          $\sfrac{2}{4}$ & $\sfrac{1}{2}$ & 0 &
          $\sfrac{2!(4-3)!}{4!}=\sfrac{1}{12}$ & 0
          \\
          $\{1,3\}$ & 1,2,5,6 & 1,5 & 
          $\sfrac{2}{4}$ & $\sfrac{1}{2}$ & 0 &
          $\sfrac{2!(4-3)!}{4!}=\sfrac{1}{12}$ & 0
          \\
          $\{2,3\}$ & 1,2,9,10 & 1,9 &
          $\sfrac{1}{4}$ & 0 & $-\sfrac{1}{4}$ &
          $\sfrac{2!(4-3)!}{4!}=\sfrac{1}{12}$ & $-\sfrac{1}{48}$
          \\
          $\{1,2,3\}$ & 1,2 & 1 & 
          $\sfrac{1}{2}$ & 0 & $-\sfrac{1}{2}$ &
          $\sfrac{3!(4-4)!}{4!}=\sfrac{1}{4}$ & $-\sfrac{1}{8}$
          \\
          \midrule[0.75pt]
          \multicolumn{7}{r}{Shapley value for feature 4 \hfill $\sv(4)~~=$} & $-0.1198$ \\
          \bottomrule[1pt]
        \end{tabular}
      \end{center}
      %%\captionof{table}{Shapley value for feature 4} \label{tab:sv4}
    \end{subfigure}
    \medskip\smallskip

    %\captionof{figure}{Shapley values for the example DT and instance $((0,0,0,0),0)$}
    \caption{Computation of Shapley values for the example DT and
      instance $((0,0,0,0),0)$. For each feature $i$, the sets to
      consider are all the sets that do not include the feature.
      For each set $\fml{S}$, we show the rows consistent with the
      values of the features in $\fml{S}$, as dictated by $\mbf{v}$.
      For example, if $\fml{S}=\{2,4\}$, then the rows of the truth
      table consistent with having features 2 and 4 assigned value 0
      are 1, 3, 9 and 11. 
      The average values are obtained by summing up the values of the
      classifier in the rows consistent with $\fml{S}$ and dividing by
      the total number of rows.
      For $\fml{S}=\{2,4\}$, only row 3 in the truth table takes value
      1, and so the average becomes $\sfrac{1}{4}$.
    }
    \label{fig:svs}
  \end{mdframed}
\end{figure*}

Besides the polynomial-time computation of Shapley values for
deterministic decomposable boolean circuits~\cite{barcelo-aaai21}, for
functions represented by truth tables, there exist polynomial-time
algorithms (on the size of the truth table) for computing the Shapley
values~\cite{hms-corr23}. This is illustrated in~\cref{fig:svs}.

\jnoteF{Highlight the key similarities between Shapley values and
  formal explanations.}

It should be noted that both formal explanations (i.e.\ AXp's and
CXp's) and Shapley values look at a prediction given a point $\mbf{v}$
in feature space, but consider the function's behavior across all
points that are consistent with the values dictated by $\mbf{v}$.
Additional similarities exist. Both formal explanations and Shapley
values look at sets of features to fix, and then analyze all the
points consistent with the fixed features. However, while formal
explanations look at some of those sets of features to fix, Shapley
values will analyze all possible subsets.

\jnoteF{Discuss what has been claimed about Shapley values, i.e. value
  0 and feature importance/non-importance!!}

The use of Shapley values in explainability have been justified by
significant claims. We illustrate some of the claims stated in earlier
work~\cite{kononenko-jmlr10}:
\begin{itemize}
\item \nohlight{\emph{``According to the 2nd axiom, if two features
  values have an identical influence on the prediction they are
  assigned contributions of equal size. The 3rd axiom says that if a
  \rhlight{\textbf{feature has no influence}} on the prediction
  \rhlight{\textbf{it is assigned a contribution of 0}}.''}}\\ 
  (Note: the axioms above refer to the axiomatic characterization of 
  Shapley values in~\cite{kononenko-jmlr10}.)
\item \nohlight{\emph{``When viewed together, these properties ensure
  that \rhlight{\textbf{any effect the features might have on the
      classifiers output will be reflected in the generated  
      contributions}}, which effectively deals with the issues of
  previous general explanation methods.''}}
\end{itemize}

Given the above, one would expect a direct correlation between a
feature's importance and the absolute value of its Shapley value.
As the rest of the paper shows, this is not the case.

\section{Feature (Ir)relevancy}
\label{sec:xpr}

Given \eqref{eq:seta} and \eqref{eq:setc}, we can aggregate the
features that occur in AXp's and CXp's:
\begin{align}
  \fml{F}_{\mbb{A}(\fml{E})}=\bigcup\nolimits_{\fml{X}\in\mbb{A}(\fml{E})}\fml{X}
  \\
  \fml{F}_{\mbb{C}(\fml{E})}=\bigcup\nolimits_{\fml{Y}\in\mbb{C}(\fml{E})}\fml{Y}
\end{align}
Moreover, MHS duality between the sets of AXp's and CXp's allows
proving that: $\fml{F}_{\mbb{A}(\fml{E})}=\fml{F}_{\mbb{C}(\fml{E})}$.
Hence, we just refer to $\fml{F}_{\mbb{A}(\fml{E})}$ as the set of
features that are contained in some AXp (or CXp).

A feature $i\in\fml{F}$ is relevant if it is contained in some AXp,
i.e.\ $i\in\fml{F}_{\mbb{A}(\fml{E})}=\fml{F}_{\mbb{C}(\fml{E})}$;
otherwise it is irrelevant,
i.e.\ $i\not\in\fml{F}_{\mbb{A}(\fml{E})}$%
\footnote{%
It should be noted that feature relevancy is tightly related with the
concept of relevancy studied in logic-based
abduction~\cite{gottlob-jacm95}.}.
We will use the predicate $\relevant(i)$ to denote that feature $i$ is
relevant, and predicate $\irrelevant(i)$ to denote that feature $i$ is
irrelevant.

Relevant and irrelevant features provide a fine-grained
characterization of feature importance, in that irrelevant features
play no role whatsoever in prediction sufficiency.
In fact, if $p\in\fml{F}$ is an irrelevant feature, then we can write: 
\begin{align}
  \forall(&\fml{X}\in\mbb{A}(\fml{E})).
  \forall(u_p\in\fml{D}_p).
  \forall(\mbf{x}\in\mbb{F}).\nonumber\\
  &\left[\bigland\nolimits_{i\in\fml{X}}(x_i=v_i)\land
  %%\nonumber\\
  (x_p=u_p)\right]\limply(\kappa(\mbf{x})=\kappa(\mbf{v}))
  %%\kappa(v_1,\ldots,v_{p-1},u_p,v_{p+1},\ldots,v_m))
\end{align}
The logic statement above clearly states that, if we fix the values of
the features identified by any AXp then, no matter the value picked for
feature $p$, the prediction is guaranteed to be $c=\kappa(\mbf{v})$.
The bottom line is that an irrelevant feature $p$ is absolutely
unimportant for the prediction, and so there is no reason to include 
it in a logic rule consistent with the instance.

\jnoteF{Invoke Occam's razor!!!}

%%\begin{example}
For the example DT, we have that $\mbb{A}(\fml{E})=\{\{2,3,4\}\}$ and
that $\mbb{C}(\fml{E})=\{\{2\},\{3\},\{4\}\}$, i.e.\ the explanation
problem has one AXp and three CXp's. (Recall that the computation of
both AXp's and CXp's is summarized in~\cref{fig:xps}.)
As expected,
$\fml{F}_{\mbb{A}(\fml{E})}=\fml{F}_{\mbb{C}(\fml{E})}=\{2,3,4\}$.
Hence, we conclude that feature 1 is irrelevant, and that features
2, 3 and 4 are relevant.
Observe that no AXp/CXp includes feature 1. For any AXp $\fml{X}$ this
means that, adding feature 1 to $\fml{X}$, when feature 1 is assigned
any value from its domain $\fml{D}_1$, would not change the
prediction.
%%\end{example}

There are a few notable reasons for why irrelevant features are not
considered in explanations. First, one can invoke Occam's razor (a
mainstay of ML~\cite{haussler-ipl87}) and argue for simplest
(i.e.\ irreducible) explanations. Second, if irreducibility of
explanations were not a requirement, then one could claim that a
prediction using all features would suffice, and that is never the
case. Third, the fact that irrelevant features can take \rhlight{any}
value in their domain without that impacting the prediction shows how
unimportant those features are.

%\section{Issues with Shapley Values for Explainability}
\section{Refuting Shapley Values for Explainability}
\label{sec:issues}

We now proceed to demonstrate that Shapley values for explainability 
can produce misleading information about feature importance, in that
the relative feature importance obtained with Shapley values disagrees
with the characterization of features in terms of (ir)relevancy.
Clearly, information about feature (ir)relevancy is obtained by a
rigorous, logic-based, analysis of the classifier, and so it captures
precisely essential information about how the classifier's prediction
depends (or not) on each of the features.

\subsection{Misleading Feature Importance} \label{ssec:misled}

Given the definition of Shapley values for explainability and of
irrelevant features, we show that Shapley values will provide
misleading information regarding relative feature importance,
concretely that an irrelevant feature can be assigned the
largest absolute Shapley value. Evidently, misleading information will
cause human decision makers to consider features that are absolutely
irrelevant for a prediction.

For the example DT, we have argued that feature 1 is irrelevant and
that features 2, 3 and 4 are relevant.
(The computation of AXp's and CXp's using a truth table is illustrated 
in~\cref{fig:xps}. \cref{sec:xpr} details how feature (ir)relevancy is
decided.)
Furthermore, from~\cref{fig:svs}, we obtain that,
\[
\forall(i\in\{2,3,4\}).|\sv(1)|>|\sv(i)|
\]
Thus, the feature with the largest absolute Shapley value is
\rhlight{irrelevant} for the prediction.

One might be tempted to argue that the sign of $\sv(1)$ differs from
the sign of $\sv(2)$, $\sv(3)$, $\sv(4)$, and that that could explain
% Obs: "that that" is correct in English, and it is on purpose!!!
the reported issue. However, the hypothetical relationship between the
sign of the Shapley values and their perceived impact of the value of
the prediction is a flawed argument, in that feature 1 plays no role
in setting the prediction to 0, but feature 1 also plays no role in
changing the value of the prediction. The results in the next section
further confirm that the sign of a feature's Shapley value bears no
direct influence on a (ir)relevancy of a feature.

\begin{comment}
%
%
\begin{table}[t]
  \begin{tabular}{cccc} \toprule
    \multicolumn{4}{c}{$\fml{S}=\emptyset$}\\\midrule
    $x_2$ & $x_3$ & $x_4$ & $\kappa(0,x_2,x_3,x_4)$ \\ \toprule
    0 & 0 & 0 & 0 \\
    0 & 0 & 1 & 1 \\
    0 & 1 & 0 & 1 \\
    0 & 1 & 1 & 0 \\
    1 & 0 & 0 & 1 \\
    1 & 0 & 1 & 0 \\
    1 & 1 & 0 & 0 \\
    1 & 1 & 1 & 0 \\
    \bottomrule
  \end{tabular}
\end{table}
%
%
\end{comment}

\begin{table*}[t]
  \renewcommand{\arraystretch}{1.25}
  \begin{tabular}{cl} \toprule
    Issue & Condition %%, $i,i_1,i_2\in\fml{F}$
    \\ \midrule[0.75pt]
    I1 &
    $\exists(i\in\fml{F}).[\irrelevant(i)\land(\sv(i)\not=0)]$
    \\
    I2 &
    $\exists(i_1,i_2\in\fml{F}).[\irrelevant(i_1)\land\relevant(i_2)\land(|\sv(i_1)|>|\sv(i_2)|)]$
    \\
    I3 &
    $\exists(i\in\fml{F}).[\relevant(i)\land(\sv(i)=0)]$
    \\
    I4 &
    $\exists(i_1,i_2\in\fml{F}).[\irrelevant(i_1)\land(\sv(i_1)\not=0)]\land[\relevant(i_2)\land(\sv(i_2)=0)]$
    \\
    I5 &
    $\exists(i\in\fml{F}).[\irrelevant(i)\land\forall(1\le{j}\le{m},j\not=i).|\sv(i)|>|\sv(j)|]$
    \\
    I6 &
    $\exists(i_1,i_2\in\fml{F}).[\irrelevant(i_1)\land\relevant(i_2)\land(\sv(i_1)\times\sv(i_2)>0)]$
    \\
    I7 &
    $\exists(i_1,i_2\in\fml{F}).[\irrelevant(i_1)\land\relevant(i_2)\land(|\sv(i_1)|>|\sv(i_2)|)\land(\sv(i_1)\times\sv(i_2)>0)]$
    \\
    \bottomrule
  \end{tabular}
  \caption{Identified potential issues with Shapley values}
  \label{tab:issues}
\end{table*}

\subsection{Issues with Shapley Values for Explainability}
\label{ssec:other}

By automating the analysis of boolean functions~\cite{hms-corr23},
we have been able to identify a number of issues with Shapley values
for explainability, all of which demonstrate that Shapley values can
provide misleading information about the relative important of
features.
The list of possible issues is summarized in~\cref{tab:issues}.
Observe that some issues imply the occurrence of other issues,
e.g.\ I4 implies I3, and I5 implies I2, among others.
Our goal is to highlight a comprehensive range of problems that the
use of Shapley values for explainability can induce, and so different
issues aim to highlight such problems.

\begin{table}[t]
  \renewcommand{\arraystretch}{1.0125}
  \begin{tabular}{lr}
    \toprule
    %Issue-related metric & Value \\
    Metric & Value \\
    \toprule
    \# of functions & 65534 \\ %65536 \\
    \# number of instances & 1048544 \\ %1048576 \\
    \midrule
    \# of I1 issues & \dbhlight{781696} \\
    \# of functions exhibiting I1 issues & 65320 \\
    \% functions exhibiting I1 issues & \rhlight{99.67}\\
    \midrule
    \# of I2 issues & \dbhlight{105184} \\
    \# of functions exhibiting I2 issues & 40448 \\
    \% functions exhibiting I2 issues & \rhlight{61.72}\\
    \midrule
    \# of I3 issues & \dbhlight{43008} \\
    \# of functions exhibiting I3 issues & 7800 \\
    \% functions exhibiting I3 issues & \rhlight{11.90}\\
    \midrule
    \# of I4 issues & \dbhlight{5728} \\
    \# of functions exhibiting I4 issues & 2592 \\
    \% functions exhibiting I4 issues & \rhlight{3.96}\\
    \midrule
    \# of I5 issues & \dbhlight{1664} \\
    \# of functions exhibiting I5 issues & 1248 \\
    \% functions exhibiting I5 issues & \rhlight{1.90}\\
    \midrule
    \# of I6 issues & \dbhlight{109632} \\
    \# of functions exhibiting I6 issues & 36064 \\
    \% functions exhibiting I6 issues & \rhlight{55.03}\\
    \midrule
    \# of I7 issues & \dbhlight{11776} \\
    \# of functions exhibiting I7 issues & 7632 \\
    \% functions exhibiting I7 issues & \rhlight{11.65}\\
    \bottomrule
  \end{tabular}
  \caption{Results over all 4-variable boolean functions.
    The two constant functions were discarded, since $\kappa$ is
    required not to be constant.} \label{tab:res}
\end{table}

By analyzing all possible boolean functions defined on four
variables,~\cref{tab:res} summarizes the percentage of functions
exhibiting the identified issues.
For each possible function, the truth table for the function serves as
the basis for the computation of all explanations, for deciding feature
(ir)relevancy, and for the computation of Shapley values.
The algorithms used are the ones sketched earlier in the paper, and
all run in polynomial-time on the size of the truth table.
For example, whereas issue I5, which is exemplified by the example DT
and instance (also, see~\cref{ssec:misled}), occurs in 1.9\% of the
functions, issues I1, I2 and I6 occur in more than 55\% of the
functions, with I1 occurring in more than 99\% of the functions.
It should be noted that the identified issues were distributed evenly
for instances where the prediction takes value 0 and instances where
the prediction takes value 1.
Moreover, it should be restated that the two constant functions were
ignored.

\subsection{Verdict \& Justification}

First, it should be plain that any of the issues described
in~\cref{tab:issues} should be perceived as problematic in terms of
assigning relative importance to features, with some issues serving to
confirm the existence of misleading relative feature importance.
This is the case with issues I2, I4, I5 and I7.
However, assigning a Shapley value of 0 to a relevant feature or
assigning non-zero Shapley value to an irrelevant feature will also
cause human decision makers to overlook important features, or to
analyze unimportant features. Such cases are also covered by the
remaining issues.

Second, and as the results of the previous two sections amply
demonstrate, the concept of Shapley values for explainability is
fundamentally flawed. Furthermore, any explainability tool whose
theoretical underpinnings are Shapley values for explainability is
also fundamentally flawed.

Third, given the similarities between the computation of abductive and
contrastive explanations and Shapley values for explainability, an
immediate question is: why do Shapley values for explainability
produce misleading measures of relative feature importance? It seems
apparent that, whereas in the original definition of Shapley values for
game theory, all coalitions are acceptable, this is not the case with
explainability, i.e.\ some sets of features should not be considered
when assigning importance to a feature. Thus, one reason that causes
Shapley values to produce misleading information is the fact that some
disallowed set of features are accounted for.

\section{Discussion} \label{sec:conc}

This paper presents a simple argument demonstrating that Shapley
values for explainability can produce misleading information regarding
relative feature importance. A number of potential issues with Shapley
values for explainability has been identified, and shown to occur
rather frequently in boolean classifiers.
%
%The results in this paper are clear in that Shapley values for
%explainability are guaranteed to provide misleading information about
%relative feature importance.
It is therefore plain that the continued use of XAI approaches based
on Shapley values (see~\cite{hms-corr23} for additional references) in
high-risk domains will inevitably cause human decision makers to
assign importance to unimportant features, and to overlook important
features. Evidently, such uses of Shapley values for explainability
are bound to have unwanted grave consequences.

If exact computation of Shapley values for explainability yields
misleading information regarding relative feature importance then,
evidently, any XAI tool that claims to approximate Shapley values for
explainability, e.g.\ SHAP and
variants~\cite{lundberg-nips17,lundberg-naturemi20,lundberg-corr22},
cannot guarantee not to produce misleading relative feature
importance.

More importantly, it should be underlined that our recent experimental
results~\cite{hms-corr23} suggest little to no correlation between
exact Shapley values and the results produced by
SHAP~\cite{lundberg-nips17}.
To put it bluntly, a flawed approximation of a flawed concept does not
offer any guarantees whatsoever regarding the rigor of that flawed
approximation at estimating feature attribution values. 
Evidently, the continued practical use of tools that approximate
Shapley values is also bound to have unwanted grave consequences.

\begin{comment}
%
In hindsight, other flaws of Shapley values for explainability should
be apparent. When the set of classes exhibits no order relation among
the different classes, then it seems clear that $\phi$ becomes
ill-defined. For example, if we target classification of animals given
a number of their attributes, the average value of the prediction
becomes meaningless. 
%
To the best of our knowledge, no solution has been proposed for this
and related problems.
%
\end{comment}

%%\jnote{Possible fixes?}

Given the demonstrated inadequacy of Shapley values for
explainability, a natural line of research is whether an alternative
metric could be devised which respects feature (ir)relevancy. Although
recent work proposed a possible metric~\cite{hms-corr23}, it is
unclear how it could be related with Shapley values for
explainability.

%\input{wip.tex}
\begin{acks}
  This work was supported by the AI Interdisciplinary Institute ANITI,
  funded by the French program ``Investing for the Future -- PIA3''
  under Grant agreement no.\ ANR-19-PI3A-0004,
  and by the H2020-ICT38 project COALA ``Cognitive Assisted agile 
  manufacturing for a Labor force supported by trustworthy Artificial
  intelligence''.
  This work was motivated in part by discussions with several colleagues
  including L.~Bertossi, A.~Ignatiev, N.~Narodytska, M.~Cooper, Y.~Izza,
  R.\ Passos, J.\ Planes and N.~Asher.
\end{acks}

%%\clearpage

% RequiredL: \usepackage{etoolbox}
%\providetoggle{mkbbl}
\newtoggle{mkbbl}
% Contents if using bibtex: "\settoggle{mkbbl}{true}"
% Contents if inputing pre-generated file: "\settoggle{mkbbl}{false}"

\settoggle{mkbbl}{false}
 % file is automatically generated

%%\cleardoublepage %% TENTATIVE, and required if bibliography starts page...
%%\addcontentsline{toc}{section}{References}
%%\vskip 0.2in
% For arxix paper production, and since arXiv does not allow for
% bibtex, we need to create a .bbl file to include upon submission
% to arXiv.
\iftoggle{mkbbl}{
  % Run bibtex, i.e. generate .bbl gile
  %\bibliographystyle{splncs04}
  \bibliographystyle{ACM-Reference-Format}
  \bibliography{refs}
}{
  % Import bibl (original .bbl) file

}
%\bibliographystyle{abbrv}
%\bibliography{refs,xtra}
%\input{wip}
%

\end{document}